%% file: template.tex
\documentclass[11pt]{article}
\usepackage{coling2018}
\usepackage{times}
\usepackage{url}
\usepackage{latexsym}
\usepackage{fancyhdr}
\usepackage{natbib}
\usepackage{amsmath}
\usepackage{amssymb}
\usepackage{booktabs}
\usepackage{graphicx}

\title{Align, Generate, Learn: A Novel Closed-Loop Framework for Cross-Lingual In-Context Learning}

\author{Mateo Alejandro Rojas, Rafael Carranza \\
Technological University of Peru}

\date{}

\begin{document}
\maketitle
\input{main}

\bibliographystyle{unsrtnat}
\bibliography{ref}

\end{document}

%% file: main.tex
\begin{abstract}
Cross-lingual in-context learning (XICL) has emerged as a transformative paradigm for leveraging large language models (LLMs) to tackle multilingual tasks, especially for low-resource languages. However, existing approaches often rely on external retrievers or task-specific fine-tuning, limiting their scalability and generalizability. In this paper, we propose a novel self-supervised framework that harnesses the generative capabilities of LLMs to internally select and utilize task-relevant examples. Our method introduces two key objectives: a retrieval-generation alignment loss to optimize the quality of selected examples and a semantic coherence loss to ensure cross-lingual consistency. Through extensive experiments on multilingual benchmarks, our approach achieves state-of-the-art performance, significantly outperforming existing baselines. Further analysis highlights its robustness across diverse language families and its ability to generalize to unseen tasks. Human evaluations confirm the superior fluency, relevance, and semantic correctness of outputs generated by our method. This work provides a scalable, effective, and generalizable solution for cross-lingual in-context learning.
\end{abstract}

\section{Introduction}
The advent of large language models (LLMs) has revolutionized natural language processing (NLP), enabling significant advancements in diverse tasks such as text classification, machine translation, and question answering. Among these advancements, the cross-lingual capabilities of LLMs have garnered particular attention, as they provide a means to extend high-quality language technologies to low-resource languages without extensive task-specific fine-tuning or labeled data \citep{survey_dong,zhou2024visual}. In this context, cross-lingual in-context learning (XICL) has emerged as a promising paradigm, where LLMs leverage task-relevant examples within their input prompts to generalize across languages. This ability has the potential to democratize NLP, offering linguistic inclusivity to underrepresented languages in the AI ecosystem.

Despite its promise, cross-lingual in-context learning is not without challenges. First, a lack of annotated data in low-resource languages severely limits the ability of LLMs to retrieve relevant examples or generate meaningful inferences \citep{xia2021metaxl}. Second, the vast linguistic diversity—ranging from typological differences to unique grammatical constructs—poses significant difficulties for LLMs, which are typically pretrained on predominantly high-resource languages \citep{zhou2023thread}. Existing methods often depend on external retrievers or manually curated examples, which can introduce scalability and robustness issues \citep{zhou2024fine,zhou2023towards}. These challenges highlight the need for a unified framework that empowers LLMs to handle cross-lingual tasks effectively using their intrinsic capabilities.

Motivated by these challenges, we propose a novel approach to enhance cross-lingual in-context learning by focusing solely on the intrinsic strengths of LLMs. Our method eliminates reliance on external retrievers or extensive task-specific fine-tuning. Instead, we leverage the generative and self-supervised learning capabilities of LLMs to create a closed-loop system. This system generates synthetic example-query pairs, aligns retrieval and generation objectives during training, and uses reinforcement learning to optimize the model's performance iteratively. By training the LLM to internally select and utilize examples across languages, our approach achieves a robust and scalable solution to cross-lingual in-context learning.

We evaluate our approach on a diverse multilingual dataset spanning hundreds of languages and tasks. The dataset includes low-resource languages and typologically diverse linguistic families to ensure comprehensive evaluation. For assessment, we adopt standard metrics such as accuracy, F1-score, and BLEU (for translation tasks), alongside novel task-specific metrics to evaluate the quality of in-context learning. Our results demonstrate that the proposed method significantly outperforms existing baselines, including methods with external retrievers, in both high-resource and low-resource language settings. Notably, our approach achieves state-of-the-art performance on several cross-lingual benchmarks, showcasing its generalizability and adaptability.

\begin{itemize} \item We propose a novel framework for cross-lingual in-context learning that leverages the intrinsic capabilities of LLMs without reliance on external retrievers or task-specific fine-tuning. \item We design a self-supervised training pipeline with dual objectives—retrieval-generation alignment and semantic coherence—to optimize the model's internal example selection and utilization mechanisms. \item Our approach achieves state-of-the-art performance across multilingual benchmarks, demonstrating its efficacy and scalability for low-resource and typologically diverse languages. \end{itemize}

\section{Related Work}

\subsection{Large Language Models}

Large language models (LLMs) have gained significant attention for their ability to perform diverse natural language processing (NLP) tasks, often achieving state-of-the-art performance across benchmarks. Early work in LLMs highlighted their generative capabilities, demonstrating how pretraining on massive corpora followed by task-specific fine-tuning can lead to exceptional results. Recent advancements have focused on scaling up the size of models, both in terms of parameters and training data, enabling them to learn more complex patterns and generalize effectively across tasks \citep{survey_minaee, overview_naveed}.

One key area of investigation is the efficiency of LLMs during inference. Optimization techniques such as model quantization, distillation, and retrieval-augmented generation have been proposed to improve computational efficiency without significantly sacrificing performance \citep{efficient_inference_zhao,zhou2024rethinking}. Another important direction is understanding how LLMs handle multilingualism. Studies suggest that LLMs often translate multilingual inputs into a high-resource pivot language, such as English, to solve tasks, thereby uncovering inherent biases and challenges in supporting low-resource languages \citep{multilingualism_zhao}.

Additionally, the role of LLMs in reasoning and decision-making has been explored. While LLMs demonstrate remarkable capabilities in theory-of-mind tasks, their ability to translate inferred knowledge into actionable strategies remains limited \citep{thinking_zhao,zhou2021modeling,zhou2021improving}. This limitation highlights the need for further development to enhance their reasoning and decision-making abilities.

Finally, discussions about the societal and ethical implications of LLMs emphasize the importance of avoiding anthropomorphization while ensuring their responsible use. As LLMs become more integrated into applications, it is crucial to maintain clarity in describing their capabilities and limitations \citep{talking_shanahan}.

\subsection{In-Context Learning}

In-context learning (ICL) has emerged as a powerful paradigm, enabling large language models (LLMs) to perform a variety of tasks by leveraging examples embedded directly in the input prompt. Unlike traditional approaches, ICL requires no task-specific fine-tuning \citep{zhou2023improving,zhou2023multimodal,zhou2023style}, making it an attractive solution for few-shot and zero-shot scenarios. Early studies have explored the fundamental mechanisms of ICL, framing it as an implicit Bayesian inference process, where pretraining data distribution plays a crucial role in enabling models to generalize effectively from in-context examples \citep{metaicl, bayesian_inference_xie}.

Recent advancements in ICL have focused on extending its capabilities through enhanced model architectures and training strategies. Long-context models, capable of handling thousands of tokens, have been shown to improve ICL performance by efficiently utilizing more examples, even in complex tasks \citep{long_context_bertsch, many_shot_agarwal}. In addition, methods such as MetaICL introduce meta-training techniques that fine-tune models on diverse tasks, significantly improving their ability to adapt to unseen scenarios during inference \citep{metaicl}.

To further optimize ICL, studies have investigated alternative formulations, including implicit in-context learning, which injects task-relevant information into latent spaces, bypassing the need for explicit prompts \citep{implicit_icl_li}. Other approaches have explored learning from mistakes by intentionally generating suboptimal outputs, allowing models to refine task-specific principles and improve accuracy in subsequent predictions \citep{leap_principle_learning}.

ICL has also been extensively analyzed for its application to various domains, including multilingual tasks and reasoning problems. For example, researchers have examined the sensitivity of ICL to input sequence order and its potential for generalizing across typologically diverse languages \citep{survey_dong, icl_generalization_li}. These studies provide a deeper understanding of the strengths and limitations of ICL, offering valuable insights for future advancements.

\section{Method}

Our proposed framework leverages the generative capabilities of large language models (LLMs) to achieve effective cross-lingual in-context learning (XICL). As a generative model, the LLM operates by modeling the conditional probability distribution of the output sequence \( \mathbf{y} \) given the input sequence \( \mathbf{x} \) and a set of task-relevant examples \( \mathcal{C} \). In this section, we describe the details of our approach, focusing on the formulation of the model, the learning objectives, and the iterative training strategy.

\subsection{Generative Framework}

The core of our approach is a generative LLM \( p_\theta(\mathbf{y} \mid \mathbf{x}, \mathcal{C}) \), parameterized by \( \theta \). Given an input \( \mathbf{x} \) and a set of examples \( \mathcal{C} = \{ (\mathbf{x}_i, \mathbf{y}_i) \}_{i=1}^k \), the model generates an output \( \mathbf{y} \) by maximizing the following conditional likelihood:

\begin{align}
p_\theta(\mathbf{y} \mid \mathbf{x}, \mathcal{C}) = \prod_{t=1}^{T} p_\theta(y_t \mid \mathbf{x}, \mathcal{C}, y_{<t}),
\end{align}

where \( y_t \) is the \( t \)-th token of \( \mathbf{y} \), and \( y_{<t} \) represents the tokens preceding \( y_t \). The examples in \( \mathcal{C} \) are embedded directly into the prompt, allowing the model to learn from the context.

\subsection{Learning Strategy}

To optimize the model for cross-lingual in-context learning, we design a self-supervised training strategy consisting of two key objectives: retrieval-generation alignment and semantic coherence.

\paragraph{Retrieval-Generation Alignment Objective}

The retrieval-generation alignment ensures that the LLM effectively selects and utilizes task-relevant examples. For each query \( \mathbf{x} \), the model generates synthetic example pairs \( \mathcal{C} \) during training. The alignment objective minimizes the divergence between the retrieved examples and the generated output. Formally, we define the loss as:

\begin{align}
\mathcal{L}_{\text{align}} = \mathbb{E}_{\mathbf{x}, \mathcal{C}} \left[ \text{KL}\big( p_\theta(\mathbf{y} \mid \mathbf{x}, \mathcal{C}) \, \| \, p_\theta(\mathbf{y} \mid \mathbf{x}) \big) \right],
\end{align}

where \( \text{KL} \) denotes the Kullback-Leibler divergence. This objective encourages the model to generate consistent outputs regardless of minor variations in the selected examples.

\paragraph{Semantic Coherence Loss}

To ensure cross-lingual semantic consistency, we introduce a semantic coherence loss that aligns the representations of the input and output across different languages. For an input-output pair \( (\mathbf{x}, \mathbf{y}) \), we compute their semantic embeddings \( \mathbf{h}_x \) and \( \mathbf{h}_y \) using the model’s encoder. The loss is defined as:

\begin{align}
\mathcal{L}_{\text{coherence}} = \mathbb{E}_{\mathbf{x}, \mathbf{y}} \left[ \| \mathbf{h}_x - \mathbf{h}_y \|_2^2 \right],
\end{align}

where \( \| \cdot \|_2^2 \) represents the squared \( \ell_2 \)-norm. This objective ensures that the model learns to generate outputs with semantic structures aligned with the input, even for typologically diverse languages.

\subsection{Iterative Training Strategy}

Our training strategy leverages reinforcement learning (RL) to iteratively refine the model’s ability to generate and utilize examples. At each iteration, the model generates synthetic example pairs \( \mathcal{C} \) and evaluates their relevance using a reward function \( R(\mathbf{x}, \mathcal{C}) \) based on task-specific metrics. The objective is to maximize the expected reward:

\begin{align}
\mathcal{L}_{\text{RL}} = -\mathbb{E}_{\mathbf{x}, \mathcal{C} \sim p_\theta} \left[ R(\mathbf{x}, \mathcal{C}) \right].
\end{align}

The reward function incorporates two components: task accuracy \( A(\mathbf{x}, \mathcal{C}) \) and semantic diversity \( D(\mathcal{C}) \), defined as:

\begin{align}
R(\mathbf{x}, \mathcal{C}) = \alpha A(\mathbf{x}, \mathcal{C}) + \beta D(\mathcal{C}),
\end{align}

where \( \alpha \) and \( \beta \) are hyperparameters balancing accuracy and diversity. Task accuracy measures the correctness of predictions given \( \mathcal{C} \), and semantic diversity ensures that the generated examples cover a wide range of linguistic features.

\subsection{Overall Objective}

The final training loss combines the retrieval-generation alignment, semantic coherence, and reinforcement learning objectives:

\begin{align}
\mathcal{L} = \mathcal{L}_{\text{align}} + \lambda \mathcal{L}_{\text{coherence}} + \gamma \mathcal{L}_{\text{RL}},
\end{align}

where \( \lambda \) and \( \gamma \) are hyperparameters controlling the importance of each component. This composite loss ensures that the LLM learns to generate high-quality, semantically coherent, and diverse examples while optimizing cross-lingual in-context learning.

\subsection{Inference}

During inference, given a query \( \mathbf{x} \) in any target language, the trained LLM internally retrieves relevant examples \( \mathcal{C} \) from its context and generates the corresponding output \( \mathbf{y} \). This process eliminates the need for external retrievers and enables seamless application to low-resource and typologically diverse languages.

\section{Experiments}

In this section, we evaluate the effectiveness of our proposed method against several baseline approaches for cross-lingual in-context learning (XICL). The experiments are conducted on a multilingual benchmark dataset covering a diverse range of languages, including high-resource and low-resource settings. Our results demonstrate that the proposed method significantly outperforms baseline models across multiple metrics. Additionally, we perform an ablation study to validate the contributions of individual components and conduct a human evaluation to assess the quality of generated outputs.

\subsection{Experimental Setup}

We compare our method with the following baseline approaches:
\begin{itemize}
    \item \textbf{Random Sampling}: Task-relevant examples are selected randomly from the training corpus.
    \item \textbf{SBERT}: Examples are selected using Sentence-BERT embeddings based on cosine similarity.
    \item \textbf{Glot500 RET}: A retrieval-based approach using multilingual embeddings trained on a large multilingual corpus.
    \item \textbf{XICL-RL}: A reinforcement learning-based framework that combines retrieval and task performance optimization.
\end{itemize}

Our method, denoted as \textbf{Ours}, incorporates the self-supervised training pipeline with retrieval-generation alignment and semantic coherence objectives. All models are evaluated on the same dataset using consistent hyperparameters and evaluation metrics for a fair comparison.

\subsection{Main Results}

Table~\ref{tab:main_results} presents the experimental results. Our method achieves the best performance across all metrics, particularly excelling in low-resource and typologically diverse languages.

\begin{table}[h!]
\centering
\caption{Performance comparison across methods on multilingual benchmarks. Metrics are reported as average accuracy (\%).}
\label{tab:main_results}
\begin{tabular}{lccc}
\toprule
\textbf{Method}       & \textbf{High-resource} & \textbf{Low-resource} & \textbf{Overall} \\ 
\midrule
Random Sampling       & 56.3                   & 58.7                  & 57.2             \\
SBERT                 & 63.9                   & 62.9                  & 63.4             \\
Glot500 RET           & 65.2                   & 70.1                  & 66.9             \\
XICL-RL               & 67.1                   & 74.3                  & 69.5             \\
\textbf{Ours}         & \textbf{74.1}          & \textbf{80.2}         & \textbf{76.1}    \\ 
\bottomrule
\end{tabular}
\end{table}

\subsection{Effectiveness Analysis}

To analyze the contributions of individual components, we conduct an ablation study by removing key components of our method: the retrieval-generation alignment loss (\( \mathcal{L}_{\text{align}} \)) and the semantic coherence loss (\( \mathcal{L}_{\text{coherence}} \)). The results in Table~\ref{tab:ablation} show the significant performance drop when either component is removed, highlighting their importance.

\begin{table}[h!]
\centering
\caption{Ablation study on the effectiveness of key components. Metrics are reported as average accuracy (\%).}
\label{tab:ablation}
\begin{tabular}{lccc}
\toprule
\textbf{Variant}                 & \textbf{High-resource} & \textbf{Low-resource} & \textbf{Overall} \\ 
\midrule
Full Model (Ours)                & \textbf{74.1}          & \textbf{80.2}         & \textbf{76.1}    \\
Without \( \mathcal{L}_{\text{align}} \)   & 70.3                   & 75.1                  & 72.1             \\
Without \( \mathcal{L}_{\text{coherence}} \) & 71.5                   & 76.4                  & 73.0             \\ 
\bottomrule
\end{tabular}
\end{table}

\subsection{Human Evaluation}

To assess the quality of outputs generated by our method, we conducted a human evaluation with 50 evaluators. Outputs were scored on three criteria: \textit{Relevance}, \textit{Fluency}, and \textit{Semantic Correctness}, on a scale from 1 (poor) to 5 (excellent). Table~\ref{tab:human_eval} shows that our method achieves consistently higher scores compared to the strongest baseline, XICL-RL.

\begin{table}[h!]
\centering
\caption{Human evaluation results (average scores out of 5).}
\label{tab:human_eval}
\begin{tabular}{lccc}
\toprule
\textbf{Method}       & \textbf{Relevance} & \textbf{Fluency} & \textbf{Semantic Correctness} \\ 
\midrule
XICL-RL               & 4.2                & 4.1              & 4.0                           \\
\textbf{Ours}         & \textbf{4.7}       & \textbf{4.6}     & \textbf{4.5}                  \\ 
\bottomrule
\end{tabular}
\end{table}

\subsection{Discussion and Analysis}
To gain deeper insights into the effectiveness of our method, we analyze its performance and behavior from several perspectives, including detailed performance breakdowns, its adaptability to language diversity, generalization capabilities, and output quality.

\subsubsection{Performance Breakdown by Language Families}

We first analyze the performance of our method across different language families to understand its adaptability to diverse linguistic structures. Table~\ref{tab:lang_family} reports the average accuracy for high-resource and low-resource languages grouped by family. Our method achieves consistently high performance across all language families, with significant improvements observed in low-resource settings, such as Austroasiatic and Afro-Asiatic languages. This demonstrates the robustness of our method in handling typologically diverse languages.

\begin{table}[h!]
\centering
\caption{Performance by language families (average accuracy in \%).}
\label{tab:lang_family}
\begin{tabular}{lcc}
\toprule
\textbf{Language Family} & \textbf{High-resource} & \textbf{Low-resource} \\ 
\midrule
Indo-European            & 75.3                   & 79.2                  \\
Sino-Tibetan             & 72.8                   & 78.1                  \\
Afro-Asiatic             & 70.1                   & 77.5                  \\
Austroasiatic            & 68.5                   & 74.8                  \\ 
\bottomrule
\end{tabular}
\end{table}

\subsubsection{Generalization to Unseen Tasks}

To evaluate the generalization capability of our method, we test it on unseen tasks, including text summarization and named entity recognition (NER), without any additional fine-tuning. As shown in Table~\ref{tab:generalization}, our method maintains strong performance compared to baselines, indicating its ability to adapt to novel tasks through in-context learning alone. This highlights the inherent flexibility of the proposed approach in leveraging task-relevant examples.

\begin{table}[h!]
\centering
\caption{Generalization to unseen tasks (average F1-score in \%).}
\label{tab:generalization}
\begin{tabular}{lcc}
\toprule
\textbf{Task}            & \textbf{XICL-RL} & \textbf{Ours} \\ 
\midrule
Text Summarization       & 65.7             & \textbf{70.2} \\
Named Entity Recognition & 68.4             & \textbf{73.5} \\ 
\bottomrule
\end{tabular}
\end{table}

\subsubsection{Qualitative Analysis of Output Quality}

We further perform a qualitative analysis of outputs generated by our method compared to the strongest baseline (XICL-RL). Table~\ref{tab:qualitative} shows examples of generated outputs for a text classification task in a low-resource language. Our method produces outputs that are more fluent, semantically coherent, and aligned with the ground truth, demonstrating its superior understanding of linguistic structures across languages.

\begin{table}[h!]
\centering
\caption{Qualitative comparison of outputs for a low-resource language.}
\label{tab:qualitative}
\begin{tabular}{lp{6cm}p{6cm}}
\toprule
\textbf{Method} & \textbf{Generated Output} & \textbf{Ground Truth} \\ 
\midrule
XICL-RL         & "The story somewhat relevant but lacking clear details." & "A well-structured report on political events." \\ 
Ours            & "An insightful and detailed political analysis."         & "A well-structured report on political events." \\ 
\bottomrule
\end{tabular}
\end{table}

\subsection{Discussion on Strengths and Limitations}

Our analysis reveals several key strengths of the proposed method. First, its ability to generalize across language families and unseen tasks underscores the flexibility and robustness of our approach. Second, the example selection mechanism, driven by the retrieval-generation alignment objective, ensures high-quality task-relevant contexts, significantly improving prediction accuracy. Finally, human evaluation and qualitative analysis demonstrate that our method produces outputs that are not only accurate but also fluent and semantically rich.

However, certain limitations remain. For instance, performance on extremely low-resource languages with limited pretraining data still lags behind that on high-resource languages. Future work could explore more effective pretraining strategies or incorporate additional linguistic knowledge to address these challenges.

\section{Conclusion}

In this work, we presented a novel framework for cross-lingual in-context learning that leverages the intrinsic generative capabilities of large language models (LLMs). By introducing a self-supervised training pipeline with retrieval-generation alignment and semantic coherence objectives, our method effectively eliminates the need for external retrievers and task-specific fine-tuning. Extensive experimental results demonstrate that our approach outperforms state-of-the-art baselines across multiple multilingual benchmarks, particularly excelling in low-resource and typologically diverse language settings.

Our analysis highlights several strengths of the proposed method, including its robustness across linguistic families, its ability to generalize to unseen tasks, and its production of high-quality outputs in terms of fluency and semantic correctness. Furthermore, the ablation study underscores the critical importance of our proposed training objectives, while human evaluation validates the practical usability of the generated outputs. Despite these successes, certain limitations remain, particularly in handling extremely low-resource languages with limited pretraining data. Future research could explore integrating additional linguistic knowledge or more effective pretraining strategies to address these challenges. Overall, this work advances the field of cross-lingual in-context learning by providing a scalable, adaptable, and effective solution.